%% file: main.tex
\documentclass[letterpaper]{article} %
\usepackage[submission]{aaai25}  %
\usepackage{times}  %
\usepackage{helvet}  %
\usepackage{courier}  %
\usepackage[hyphens]{url}  %
\usepackage{graphicx} %
\def\UrlFont{\rm}  %
\usepackage{natbib}  %
\usepackage{caption} %
\usepackage{algorithm}
\usepackage{algorithmic}

\usepackage{newfloat}
\usepackage{listings}
\DeclareCaptionStyle{ruled}{labelfont=normalfont,labelsep=colon,strut=off} %
\floatstyle{ruled}
\newfloat{listing}{tb}{lst}{}
\floatname{listing}{Listing}
\title{\LARGE Story3D-Agent: Exploring 3D Storytelling Visualization\\ with Large Language Models\\
\vspace{+2mm}

\Large Yuzhou Huang$^{1^{\#}}$ 
\hspace{3pt} Yiran Qin$^{1}$ 
\hspace{3pt} Shunlin Lu$^{1}$ 
\hspace{3pt} Xintao Wang$^{2}$$^\dagger$ \\
\hspace{3pt} Rui Huang$^{1}$ 
\hspace{3pt} Ying Shan$^{3}$
\hspace{3pt} Ruimao Zhang$^{1}$$^\dagger$ 

\vspace{+1.5mm}
\small$^1$The Chinese University of Hong Kong, Shenzhen (CUHK-SZ) \hspace{5pt}
\small$^2$Kuaishou Technology \hspace{5pt} \small$^3$ARC Lab, Tencent PCG \hspace{5pt} \\

\vspace{+1mm}
\small$^\#$ Work done during the internship at ARC\hspace{5pt} 
\small$^\dagger$ Corresponding author\hspace{5pt}
\vspace{+1.5mm}
}

\author {
    Author Name
}
\affiliations{
    Affiliation\\
    Affiliation Line 2
}

\usepackage{bibentry}

\usepackage{multirow}
\usepackage{booktabs}
\usepackage{amsmath}  
\usepackage{amssymb}  
\usepackage{mathrsfs} 
\usepackage{cleveref}
\usepackage{array}
\usepackage{float}

\def\eg{\emph{e.g.}}

\usepackage[dvipsnames]{xcolor}

\begin{document}

\twocolumn[{%
\renewcommand\twocolumn[1][]{#1}
\maketitle
\vspace{-12mm}
\begin{center}
    \centering
    \captionsetup{type=figure}
    \hspace*{-4mm}\includegraphics[width=0.9\linewidth]{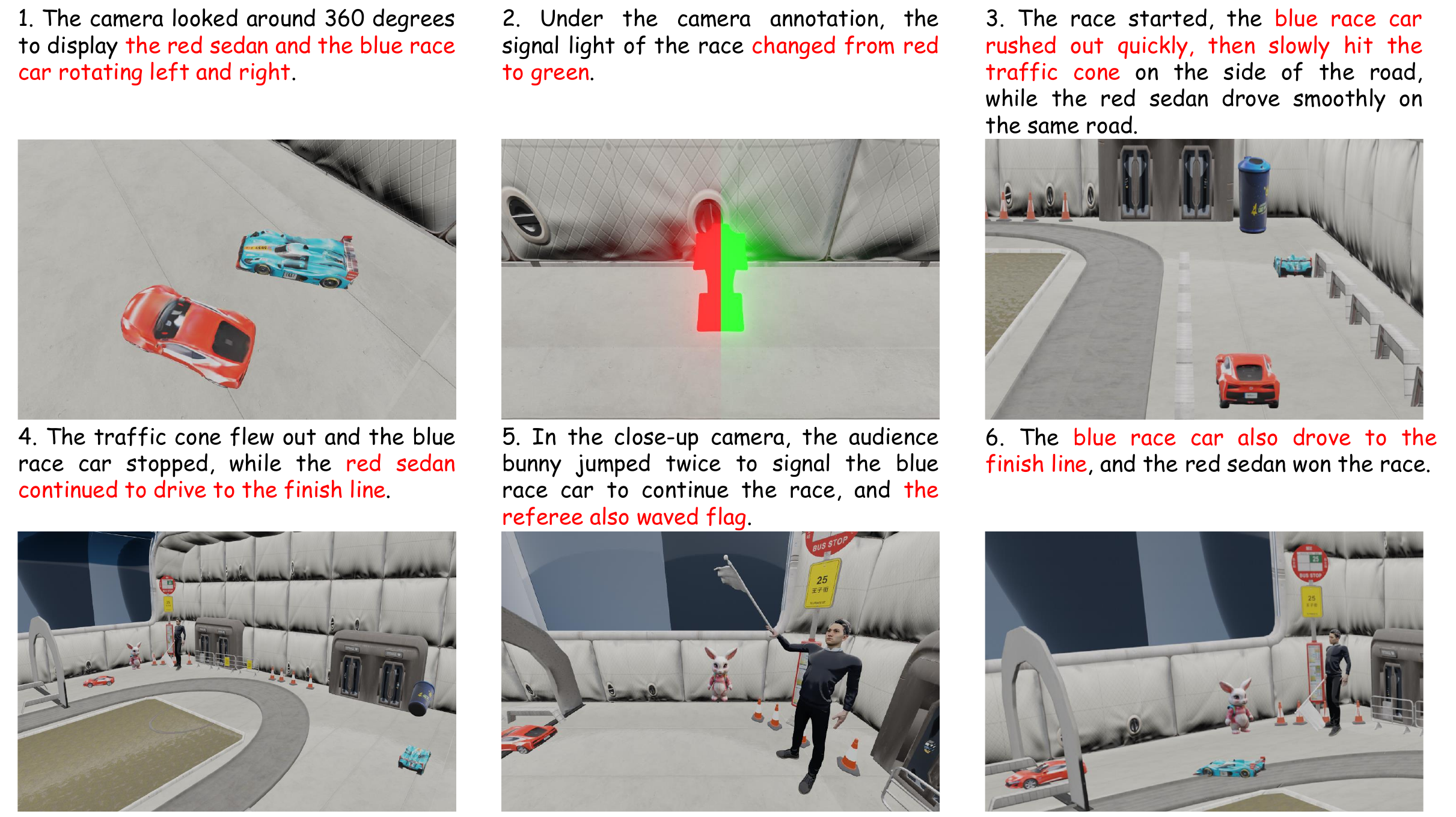}
    \vspace{-2mm}
    \captionof{figure}{
    We present \textbf{Story3D-Agent}, an innovative LLM-agents system designed for 3D storytelling visualization.
    The primary objective of the LLM-agents system is to adeptly transform a provided narrative into a corresponding 3D visualization.
    In this figure, we illustrate the narrative titled \textit{Race Day}, represented as a 3D-rendered representation.}

    \label{fig:Teasor1}
\end{center}%
}]

\input{RealSec1/0_abstract_v1}

\input{RealSec1/1_introduction_v1}
\input{RealSec1/2_related_v1}

\input{RealSec1/3_method_v1}

\input{RealSec1/4_experiment_v1}

\input{RealSec1/5_conclusion_v1}

\bibliography{main}

\clearpage
\input{RealSec1/X_Supp_v1}

\end{document}

%% file: RealSec1/0_abstract_v1.tex
\begin{abstract}

Traditional visual storytelling is complex, requiring specialized knowledge and substantial resources, yet often constrained by human creativity and creation precision.
While Large Language Models (LLMs) enhance visual storytelling, current approaches often limit themselves to 2D visuals or oversimplify stories through motion synthesis and behavioral simulation, failing to create comprehensive, multi-dimensional narratives.
To this end, we present \textbf{Story3D-Agent}, a pioneering approach that leverages the capabilities of LLMs to transform provided narratives into 3D-rendered visualizations. 
By integrating procedural modeling, our approach enables precise control over multi-character actions and motions, as well as diverse decorative elements, ensuring the long-range and dynamic 3D representation.
Furthermore, our method supports narrative extension through logical reasoning, ensuring that generated content remains consistent with existing conditions.
We have thoroughly evaluated our Story3D-Agent to validate its effectiveness, offering a basic framework to advance 3D story representation.

\end{abstract}

%% file: RealSec1/1_introduction_v1.tex
\section{Introduction}
\label{sec:intro}

Early professional guidebooks of visual content creation~\cite{lankshear2010diy, chong2007digital} emphasize that the core values, skills, and knowledge in question are all intrinsically linked to a singular concept: telling a story.
Traditional visual story creation process is complex, requiring professional knowledge and substantial resources, while always being limited by human creativity and creation precision, posing significant barriers to its evolution.

The technological advancements in Large Language Models (LLMs)~\cite{radford2018improving, radford2019language, brown2020language, achiam2023gpt, touvron2023llama, touvron2023llama2} with enhanced comprehension, planning, and reasoning abilities have significantly impacted visual storytelling. Recent studies have utilized LLMs to reason image layouts, assist visualization to videos, or empower 3D characters to simulate human psychological processes and interact through synthesized body movements~\cite{gong2023talecrafter, wang2023autostory, he2023animate, zhuang2024vlogger, cai2023digital}.

Nonetheless, depending solely on images or videos for visualization undermines story coherence and visual effects consistency, struggling to yield satisfactory outcomes. Additionally, exclusively focusing on human-centric motion synthesis and behavioral simulation proves inadequate for comprehensive story representation.
We believe that effective visual story creation necessitates maintaining long-term narrative coherence and dynamic visualization consistency in 3D format. 
Simultaneously, to facilitate superior visual effects and enhance diversity and creativity, we have incorporated numerous requisite components including: 1) diverse actions for 3D characters (\eg, curved turning by car). 2) human motions for humanoid characters (\eg, person walking). 3) various decorative elements for visual effects (\eg, camera perspective switching and environment lighting changes).

We harness the powerful LLMs to dispatch these components. Utilizing LLMs for 3D storytelling visualization is non-trivial due to a lack of inherent 3D world understanding and the difficulty in generating accurate executable codes. Procedural modeling, however, provides a way for LLMs to execute designed functions and make decisions at the cognitive level, hence effectively addressing these challenges in 3D task modeling.
In this paper, we introduce \textbf{Story3D-Agent}, a novel LLM-agents system designed for 3D storytelling visualization, whose primary objective is to convert a given narrative, which includes multiple customized characters and reference objects, into a 3D-rendered representation. 
We leverage the sophisticated capabilities of LLMs to achieve: 1) Simultaneous and precise control of actions and motions for multiple characters and various decorative elements, aligning them with the given narrative. 
2) Strict adherence of all story segments to the predetermined storyline to maintain narrative coherence and content consistency in the results of long-range and dynamic 3D visual representation. 
The aforementioned problems can be addressed by employing LLM-agents to conduct 3D storytelling via procedural modeling.

To better facilitate our target, we adhere to the traditional visual creation pipeline. Initially, we segment the given story into multiple fragments, with each fragment representing an event window. 
We further divide each event window into two phases: 
\textbf{1) Decision-making}, which involves proposed director, action, motion, and decoration agents tasked with scheduling the fragment duration, as well as coordinating actions and motions for multiple characters and decorative elements in alignment with the storyline.
\textbf{2) Textual Self-check}, which necessitates the aforementioned agents to conduct self-reflection and correction on their provided outputs, the process holds until it validates the accuracy of each output result.
Once all event windows are completed, \textbf{Execution} converts all previous outputs into executable codes on Blender\footnote{https://www.blender.org/} software. This approach facilitates the generation of 3D-rendered results that strictly adhere to the narrative timeline.
Finally, in the \textbf{Visual Self-check} stage, either human reviewers or Vision-Language Models (VLMs) could analyze the generated 3D visualizations and provide feedback, guiding the system to make necessary adjustments.

Moreover, to augment the scope for creativity in story creation, our Story3D-Agent is furnished with narrative extending capacity that safeguards continuity with the preceding narrative. 
We authorize LLMs to generate new story employing rigorous logical reasoning and reflect the rationale behind the designed contents, which assist more precise continuation. 

We have carried out extensive experiments to substantiate the efficacy of our proposed Story3D-Agent. In conclusion, we offer the following contributions:
\begin{itemize}
    \item We present the Story3D-Agent, a novel LLM-agents framework designed to utilize LLMs in exploring 3D storytelling visualization, and we aspire that our exploration aligns with the core values of visual content creation, extends creative ideas, and accelerates 3D content construction. 
    \item Our Story3D-Agent effectively resolves complexities of 3D storytelling, including narrative-aligned multi-character controls and decorative elements, and strict adherence to the predetermined story, ensuring narrative coherence and content consistency in 3D modeling.
    \item Our proposed Story3D-Agent possesses the capability to generate new stories that are contextually consistent with the provided narrative and condition, which could further promote its practicality.

\end{itemize}

%% file: RealSec1/2_related_v1.tex
\section{Related Work}
\label{sec:related work}

\subsection{LLM Agents}
LLM-agents have recently emerged as a novel class of artificial intelligence systems, exhibiting capabilities such as profiling, memory, planning, and decision-making~\cite{wang2023survey}. These agents have been extensively explored across various domains, including visual reasoning~\cite{gupta2023visual, suris2023vipergpt}, mathematical problem-solving~\cite{wei2022chain, imani2023mathprompter}, human behavior simulation~\cite{park2023generative}, gaming for embodied tasks~\cite{wang2023voyager, zhu2023ghost, qin2023mp5, zhou2024minedreamer}, and robotic navigation~\cite{huang2022inner}.

In visual content generation, LLM-agents have demonstrated significant potential across various modalities. For instance, they have been utilized for generating precise 2D images from complex textual instructions, facilitating consistent multi-scene video production, and simulating autonomous 3D characters capable of engaging in social interactions and personal expressions~\cite{cho2023visual, wu2023self, lin2023videodirectorgpt, cai2023digital}.
Moreover, traditional 3D content generation often grapples with issues stemming from the size of datasets and model performance. To resolve these limitations, LLM-agents have increasingly been utilized such as the generation of 3D scenes, as well as mesh creation and editing~\cite{sun20233d, raistrick2023infinite, yamada2024l3go, yang2023holodeck}. The use of these joint techniques offers a promising pathway to improve and streamline the 3D content creation process. In contrast, our work introduces a novel 3D storytelling visualization framework based on LLM-agents system, focusing specifically on long-range and dynamic 3D visual effects.

\begin{table*}[h]
\centering
\resizebox{0.88\textwidth}{!}{%
\begin{tabular}{|c|c|c|c|c|c|c|}
\hline
\textbf{Action} & Major Category & Sub-action & \textbf{Motion} & Category & \textbf{Decoration} & Category  \\
\hline
1 & special action & do nothing & 1 & special motion & 1 & switching camera perspective \\
2 & straight line movement & constant/variable speed movement & 2 & trajectory-based motion & 2 & light illumination \\
3 & curved movement & bezier/S-/B-curve movement & 3 & human-object interaction & 3 & particle floc \\
4 & jumping motion & jump in place/forward & 4 & human-scene interaction & 4 & beaming material \\
5 & impact motion & fall down, knocked down/away & 5 & physics-based motion & 5 & fireworks \\
6 & state recovery action & stand up, landing from the sky & & & 6 & sun light adjustment \\
7 & demonstration action & rotate in place, drifting & & & 7 & camera ring shot \\

\hline
\end{tabular}%
}
\vspace{-3mm}
\caption{Pre-defined action, motion and decoration function libraries in Blender.}
\vspace{-5mm}
\label{table:pre_defined_functions}
\end{table*}

\subsection{LLM Assists AIGC}

The open-source LLaMA framework~\cite{touvron2023llama, touvron2023llama2, vicuna2023} has significantly enhanced generative vision tasks. The pioneering work GILL~\cite{koh2023generating} bridges the gap between Multimodal Large Language Models (MLLMs)~\cite{liu2023visual, zhu2023minigpt} and diffusion models~\cite{rombach2022high}, expanding multimodal capabilities such as image retrieval, novel image generation, and multimodal dialogue. Some studies~\cite{ge2023planting, ge2023making, sun2023generative, feng2023ranni} advance text-to-image generation using LLMs, while works including~\cite{fu2023mgie, huang2023smartedit} further explore instruction-driven image editing. LLMs also play a crucial role in 3D generative tasks. For example, GALA3D~\cite{zhou2024gala3d} focuses on generating complex 3D scenes with multiple objects, and GaussianEditor~\cite{fang2023gaussianeditor} facilitates 3D scene editing.

In storytelling, LLMs have driven the creation of innovative applications. TaleCrafter~\cite{gong2023talecrafter} and Autostory~\cite{wang2023autostory} enable interactive 2D story visualization, managing multiple characters and allowing for layout and structural modifications. Animate-A-Story~\cite{he2023animate} and Vlogger~\cite{zhuang2024vlogger} extend storytelling to videos. The former assimilates coherent stories from preexisting video clips by personalizing their appearances in alignment with the story's demand, while the latter approach behaves akin to a director, parceling lengthy vlog productions into controllable stages. This strategy not only simplifies the process but also ensures each segment contributes effectively to the entire narrative.

%% file: RealSec1/3_method_v1.tex
\section{Method}
\label{sec:Method}

\subsection{3D Storytelling Visualization Formulation}
\label{sub:formulation}
Our aim is to empower the capabilities of LLM-agents in executing 3D storytelling visualization. The primary objective of this task is to transform a given narrative into a 3D-rendered representation, involving multiple customized characters and reference objects. Within the narrative, characters are required to perform corresponding actions and motions, supplemented by additional decorative elements to enhance the appeal and vivacity of generation results. 
The challenges associated with this task are twofold: 1) Simultaneous and precise control of actions and motions for multi-characters and various decorative elements to align with the given narrative. 2) To maintain narrative coherence and content consistency in long-range and dynamic 3D modeling, it is crucial that all story segments rigorously conform to the predetermined story timeline. 
The illustration of the story titled \textit{Race Day} is presented in Figure.~\ref{fig:Teasor1}, serving as an exemplar of 3D story visualization. Subsequently, we will introduce the meticulously designed Story3D-Agent.

\subsection{Preliminary of Procedural Modeling}
\label{sub:preliminary}
To facilitate the target, procedural modeling is a good method that could effectively address the challenges faced by LLMs in 3D task modeling. 
We aim at incorporating numerous requisite elements that contribute to superior visual effects and foster expansive creativity. Human motion is a pivotal aspect of the traditional animation pipeline. Creating a compelling visual story mandates the consideration of humanoid characters that exhibit natural and expressive motions to depict vibrant narrative content. Consequently, we construct an action library for 3D entities and a decoration library for narrative embellishment. In addition, we incorporate human motion generation~\cite{shafir2023human, liu2024programmable} in the manner of dispatching functionalities. These elements ensure rich and engaging storytelling effects.
Each library comprises numerous pre-defined Python functions (see Table.~\ref{table:pre_defined_functions}). Initially, we endowed LLMs with the ability to exploit three predefined libraries for visualization execution via procedural modeling. To elucidate the roles of introduced director, action, motion, and decoration agents, we provide the subsequent key definitions:

$\bigstar$ $L$(Library): This systematically introduces the objectives and functionalities included in the pre-defined library.

$\bigstar$ $F$(Function): This illustrates the functionalities of the functions contained in the pre-defined library.

$\bigstar$ $V$(Variable): This provides explanations of function variables.

$\bigstar$ $I$(In-context): This provides usage examples, demonstrating how to use provided inputs to reason out the results, which are presented in in-context format.

Furthermore, we introduce a hierarchical structure design for the construction of the action function library. Actions are initially divided into several major categories, each containing different sub-actions. and sub-actions can then be executed The major category is defined as follows:

$\bigstar$ $C$(major Category): This introduces the functionalities of major action categories in the action library.

By equipping LLMs with these essential concepts, we could empower them to utilize their generative competencies in procedural modeling, facilitating the execution of 3D story visualization.

\begin{figure*}
    \centering
    \includegraphics[width=1.0\linewidth]{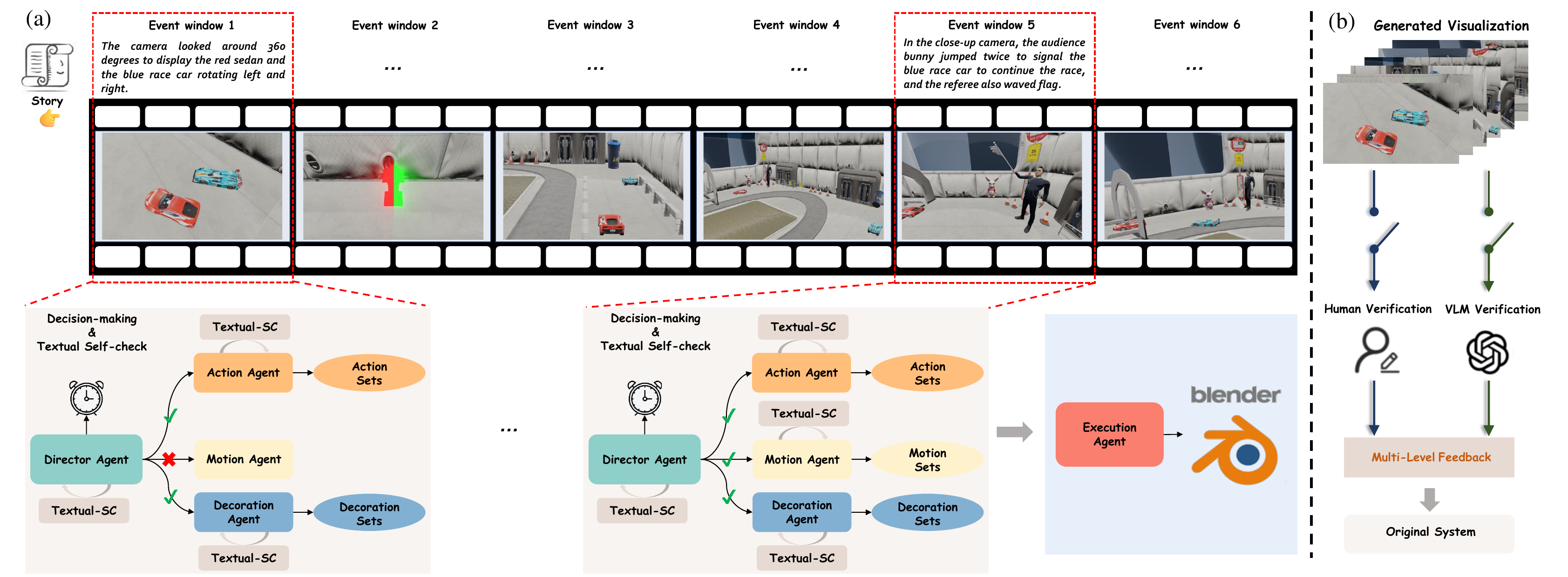}
    \vspace{-6.5mm}
    \caption{Overviews of proposed (a) \textbf{Story3D-Agent} and (b) Visual Self-check workflow. Our method involves dividing a story into multiple parts, each serving as an event window. Using LLMs, we independently determine the corresponding storyline for each clip. These determinations are then compiled for the overall story model. Further, the accuracy of the system's determinations is improved by a multi-dimensional error correction mechanism.} 

    \vspace{-5.5mm}
    \label{fig:framework}
\end{figure*}

\subsection{Framework of Story3D-Agent}
\label{sub:framework}
Figure.~\ref{fig:framework} illustrates the framework of Story3D-Agent. The approach to construct 3D story fundamentally deviates from previous LLM-agents assisted 3D modeling tasks~\cite{yamada2024l3go, yang2023holodeck, sun20233d}, oriented primarily towards long-range content consistency and dynamic visual modeling anchored to temporal timelines. We initially divide a given narrative into multiple segments based on pasring results of LLMs, each representing an event window. We anticipate that LLM-agents will firstly accomplish the determination of the corresponding storyline within each event window independently, afterwards summarizing all determinations for the overall story modeling. Multi-dimensional error correction mechanism also enhances the accuracy of system determinations. We posit that this modeling process aligns with the standard procedure of 3D storytelling visualization.

Therefore, each event window undergoes two primary processes: \textbf{1) Decision-making} process, which begins with the director agent determining which library or libraries should be dispatched and the duration required for the current segment. It then schedules actions and motions for characters and decorative elements that align with the storyline in the corresponding library, utilizing action, motion and decoration agents, respectively. 
\textbf{2) Textual Self-check} process, which involves self-reflection and correction. The director agent must assess the appropriateness of the library dispatch, while the action, motion and decoration agents must evaluate whether the selected execution functions align with the story plot. This process continues until the textual self-check mechanism verifies the accuracy of each output result. 
Once all event windows have completed the aforementioned processes, \textbf{Execution} process commences. This process receives the outputs from the action, motion and decoration agents and automatically converts them into executable codes. The software then renders all clips strictly according to the given story timeline.
\textbf{Visual Self-check} process involves human reviewers or Vision-Language Models (VLMs) analyzing and evaluating the generated visual content. The feedback is then fed back into the original system to refine the determinations for 3D storytelling visualization.

\begin{figure*}
    \centering
    \includegraphics[width=0.78\linewidth]{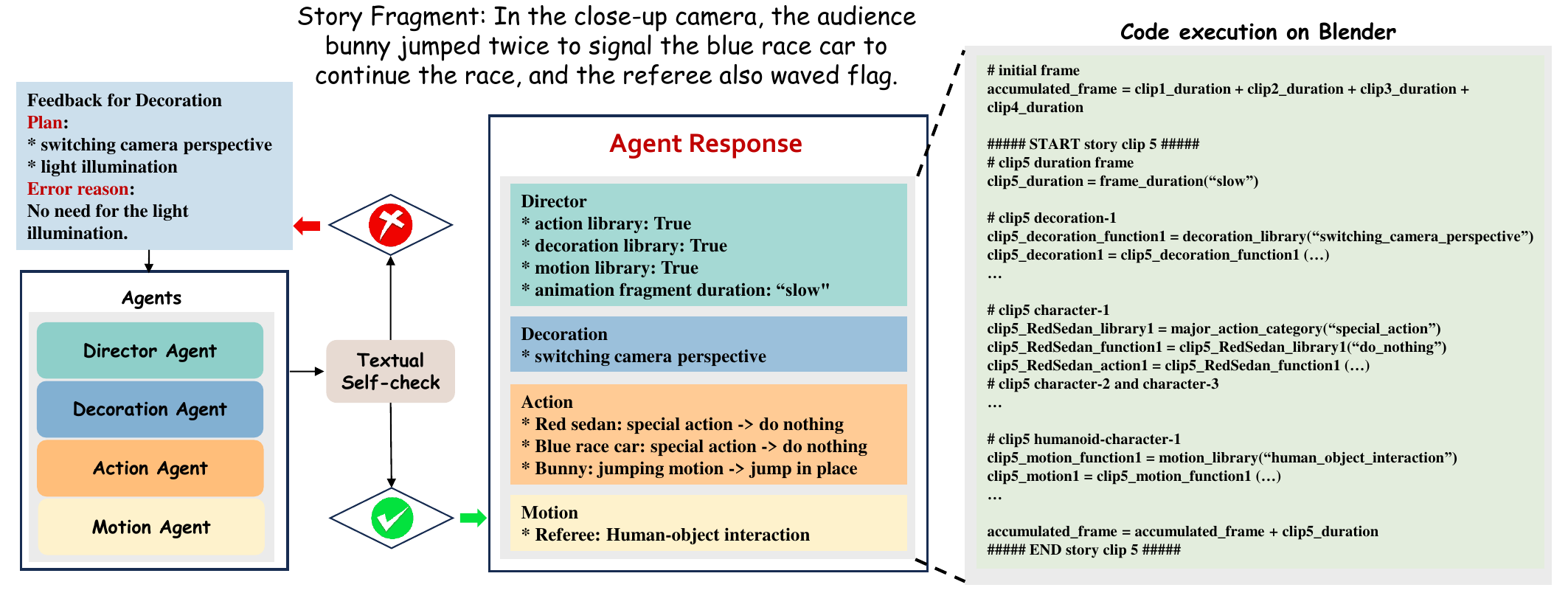}
    \vspace{-3mm}
    \caption{The director, action, motion, and decoration agents are required to initially produce their respective outputs. Subsequently, these outputs are evaluated by the textual self-check mechanism. This mechanism not only confirms the correct responses but also initiates self-reflection and correction for potential errors. The process continues until all outputs are deemed error-free, at which point the determination process within the current event window is concluded.}

    \vspace{-3mm}
    \label{fig:casestudy}
\end{figure*}

\subsubsection{Decision-making Process.}
\label{subsub:decision}
Upon receiving the fragment description, the director agent determines the appropriate library or libraries for implementation and analyzes the duration of the current fragment. The communication flow for director agent is shown as follows, and the example will be illustrated in the supplementary material:

\fbox{
\parbox{0.43\textwidth}{
\small
\textbf{---LLM system:} \textit{You are 3D story director, and you need to plan how to use given Python function libraries to create 3D story based on descriptions of fragments. You are provided three Python function libraries, namely action, motion and decoration function libraries, with detailed introduction} $L_{action}$, $L_{motion}$ \textit{and} $L_{decoration}$. \textit{Learning their functionalities that assists your analysis. Besides, give out the duration of each fragment, and choose four different types of duration including "fast", "moderate", "slow" and "emphasis". Some in-context examples} $I_{director}$ \textit{are provided to teach you how to response.}

\textbf{---LLM input:} \textit{story fragment description.}

\textbf{---Director agent response:} \textit{appropriate library or libraries and fragment duration.}
}}

\vspace{+2mm}

Subsequently, the action, motion and decoration agents select functions that align with the storyline from their respective libraries. The motion and decoration agent directly provides all scheduling functions, while the action agent conducts dispatches in a hierarchical manner, first considering the primary category of action, followed by sub-action. The communication flows for action, motion and decoration agents are shown as follows. Notice that since the procedural modeling methods employed by motion and decoration agents are identical, we have chosen to them in the ensuing workflow, and the examples for action, motion and decoration agents will be illustrated in the supplementary material:

\vspace{+2mm}
\fbox{
\parbox{0.43\textwidth}{
\small
\textbf{---LLM system:} \textit{You can execute actions for characters corresponding to descriptions of fragments based on the detailed introduction of action library} $L_{action}$, \textit{which contains several major action categories} $C_{action}$, \textit{and each major category contains one or several sub-actions} $F_{action}$. \textit{You should learn how to use functions with variables explanation} $V_{action}$. \textit{Some in-context examples} $I_{action}$ \textit{are provided to teach you how to response.}

\textbf{---LLM input:} \textit{story fragment description.}

\textbf{---Action agent response:} \textit{appropriate primary category of action and sub-action for each character.}
}}

\vspace{+2mm}
\fbox{
\parbox{0.43\textwidth}{
\small
\textbf{---LLM system:} \textit{You can execute motions for humanoid characters/decorations corresponding to descriptions of fragments based on the detailed introduction of motion/decoration library} $L_{motion}$/$L_{decoration}$, \textit{which contains several motion/decoration categories} $F_{motion}$/$F_{decoration}$. \textit{You should learn how to use functions with variables explanation} $V_{motion}$/$V_{decoration}$. \textit{You need to find out the corresponding functions to conduct the motions/decorative elements. Some in-context examples} $I_{motion}$/$I_{decoration}$ \textit{are provided to teach you how to respond in the format.}

\textbf{---LLM input:} \textit{story fragment description.}

\textbf{---Motion/Decoration agent response:} \textit{appropriate motions/decorative elements.}
}}

\vspace{+2mm}

\noindent {\footnotesize\textbf{Why do we need a director agent? }}
In each event window, the director agent is utilized to initially analyze the description from a comprehensive perspective. This allows for the simultaneous determination of which library or libraries should be dispatched to implement the story, thereby reducing the need to schedule the action, motion and decoration agents each time. It also enables determination of the duration time that suits the plot.

\noindent {\footnotesize\textbf{Why do we separate the action, motion and decoration agents rather than merging them? }}
The functionalities of the action, motion and decoration libraries have a clear semantic gap, leading to distinct impacts on story. Consequently, the content analyzed by the three agents differs, prompting us to design three separate agents to independently perform their respective functions.

\noindent {\footnotesize\textbf{Why do we design the action agent with a hierarchical dispatch mode? }}
The hierarchical dispatch mode is anthropomorphic, mirroring the thought process of creators. In visual content creation, the type of movement for the character is often chosen first (e.g., curved movement for a turning car), which is then concretized to fit the story's development (e.g., specialized bezier curve movement for driving mode). This anthropomorphic thinking mode led us to design the dispatch method for the action agent in a hierarchical manner.

\subsubsection{Agents in Decision-making Process with Textual Self-check.}
\label{subsub:self_check}
The textual self-check process involves conducting self-reflection and correction on the output results of the aforementioned four agents during the decision-making process. If the output results are deemed incorrect, feedback is provided on how to rectify them. This feedback is then used as an additional input prompt to re-run the original agent. Consequently, the decision-making and textual self-check processes continue until all output results are confirmed to be error-free. The additional reasonable prompts given by the mechanism could largely increase the truth agents' responses. We will demonstrate the examples of the processes in the supplementary material.

\subsubsection{Execution Process.}
\label{subsub:execution}
Once decision-making and textual self-check processes for each event window are completed, the execution agent assumes control of the visualization. Its primary role is to automatically integrate the signals provided by the action, motion and decoration agents, as well as time duration into executable codes, which can then be implemented on the software. This execution process is designed to adaptively accommodate a variable number of characters, reference objects and fragments in the given story and benefits long-range and dynamic temporal modeling scenarios, thereby significantly reducing the human workload and benefiting actual visualization production.

\subsubsection{Case Study.}
Figure.~\ref{fig:casestudy} as an example to elucidate the decision-making and textual self-check processes. In this fragment, director agent dispatches all three libraries, as well as determines a relatively long duration (\eg, slow). Subsequently, the outputs of the director, action and motion agents are positively confirmed, while the textual self-check within the decoration agent initiates feedback for potential inaccuracy, thereby necessitating the self-reflection first to give out the suggestion, and re-execution with additional prompting to correct the mistake. Ultimately, once the re-generated response is deemed correct (\eg, remove light illumination), the next event window is initiated.

\begin{figure*}
    \centering
    \hspace*{-5mm}\includegraphics[width=1.05\linewidth]{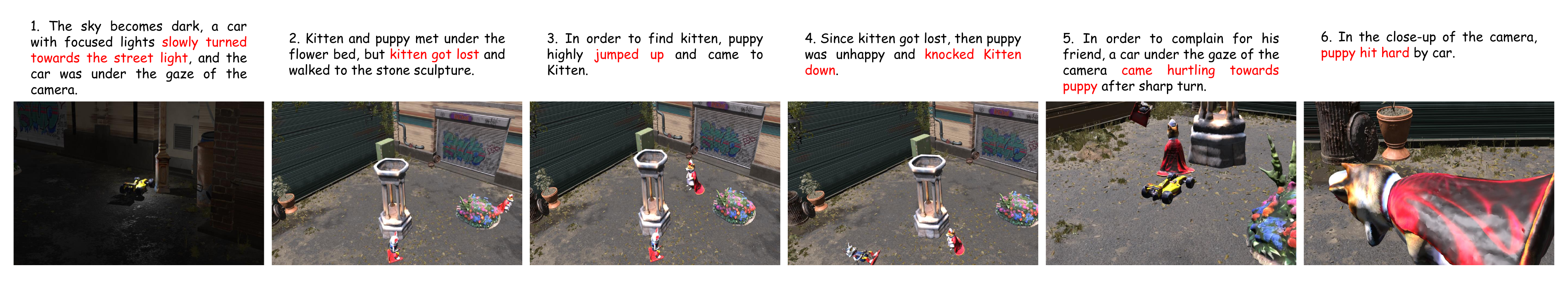}
    \vspace{-8mm}
    \caption{We depict another vivid narrative titled \textit{Friendship} provided by our \textbf{Story3D-Agent}, which narrates the story about the univeral theme of friendship in a pre-defined in garden-like scene.}

    \vspace{-6mm}
    \label{fig:Friendship3}
\end{figure*}

\subsubsection{Visual Self-check.}
Visual Self-check process comprises the analysis and evaluation of generated visual content either by human reviewers or Vision-Language Models (VLMs). 
The right part of Figure.~\ref{fig:framework} demonstrates its workflow.
Different from textual counterparts, it determines the generated contents in vision level, once it confirms the visual results, the overall system finishes the work, or it will give out additional prompt to re-run the original system.

\vspace{-3mm}
\subsection{Story Continuation}
\label{sub:continuation}
Additionally, we investigate the capability of LLMs to extend story continuation in storytelling. When generating new narratives, the created contents must maintain consistency and relevance to the previous materials. We posit that this capability offers a crucial enhancement to the demands of actual visual content generation. 
To fulfill this requirement, we need to: 1) Ensure that the newly created story can be executed by the pre-defined three function libraries. 2) Define the developmental context of the new narrative through specific conditions (such as comedy or tragedy). 3) Require that LLMs exhibit rigorous logical reasoning when crafting new stories, thereby facilitating a more seamless integration of the contextual status and content of the supplied narratives.
The communication flow is shown as follows, more examples will be illustrated in the supplementary material:

\vspace{+2mm}
\fbox{
\parbox{0.43\textwidth}{
\small
\textbf{---LLM system:} \textit{You can create 3D story, and you need to continue creating new story based on existing one. The new story you create must be implemented by the provided three Python function libraries with detailed introduction} $L_{action}$, $L_{motion}$ \textit{and} $L_{decoration}$. \textit{Some in-context examples} $I_{continuation}$ \textit{are provided to teach you how to response.}

\textbf{---LLM input:} \textit{provided story and continuing conditions.}

\textbf{---Story continuation response:} \textit{appropriate new story with logical reasoning.}
}}

%% file: RealSec1/4_experiment_v1.tex
\section{Experiments}

\vspace{-1mm}

\subsection{Experimental Settings}
\subsubsection{Blender Prerequisite Knowledge.} 
We employ the widely used 3D modeling software Blender, as our visualization execution platform. Through preliminary learning, we have acquired a certain level of proficiency in Blender's Python interface and have pre-defined three function libraries (refer to Table.~\ref{table:pre_defined_functions}). These libraries provide substantial support for our target. The modeling process incorporates the entities of the story, specifically, characters, humanoids and reference objects. We will introduce these three Python libraries with detailed functionalities in the supplementary material.

\subsubsection{3D assets acquisition.}
We utilize TripoAI\footnote{https://www.tripo3d.ai/} (a text-to-3D platform) to create 3D objects, and Mixamo\footnote{https://www.mixamo.com/} (a 3D character animation platform) for textures of humanoid characters. This approach enables designers to freely customize 3D characters and reference objects for their story. Additionally, we select 3D scene assets from open source community. Once all 3D entities are pre-selected, designers can input the narrative and execute the 3D visualization in Blender.

\subsubsection{Base model selection.}
We preliminary choose GPT-4~\cite{achiam2023gpt} as the base language model for our LLM-agents system. For the visual self-check process, we select GPT-4o~\cite{OpenAI2024} with powerful visual understanding ability, which additionally participates in assisting our evaluation.

\subsubsection{Evaluation Criteria.} 
We build three distinct criteria to evaluate our work. 
1) The first criterion assesses whether each rendered segment accurately aligns with the given story. Specifically, we employ twenty evaluators to manually assess the results. The final evaluation, termed as Instruction-Alignment (Ins-Align), is derived from the average of all these assessments. 
2) The second criterion is the CLIP score~\cite{radford2021learning}, which is computed by determining the similarities between the rendered results and the provided narrative.
3) The third evaluative criterion employs the visual understanding capabilities of GPT-4o~\cite{OpenAI2024} to caption rendered results, and we assess the discrepancy between the generated captions and the original story. We take advantage of conventional Natural Language Generation (NLG) metrics ROUGE-L~\cite{lin2004rouge}. We additionally incorporate the semantic similarity score, utilizing the BERT-base~\cite{devlin2018bert} model to attain embeddings and compute similarity. This measurement ensures that we depend on both vocabulary-level and semantic-level scores.
We will introduce more details of our evaluation in the supplementary material.

\vspace{-1mm}
\subsection{More Visualization Results}
In Figure.~\ref{fig:Friendship3}, we illustrate an additional narrative entitled \textit{Friendship} depicted through our method.
The visual narrative presented vividly features characters including kitten, puppy, and car. This story pre-defined in a garden-like scene, is constructed around the universal theme of friendship.

\vspace{-2mm}
\begin{figure*}
    \centering
    \includegraphics[width=1.0\linewidth]{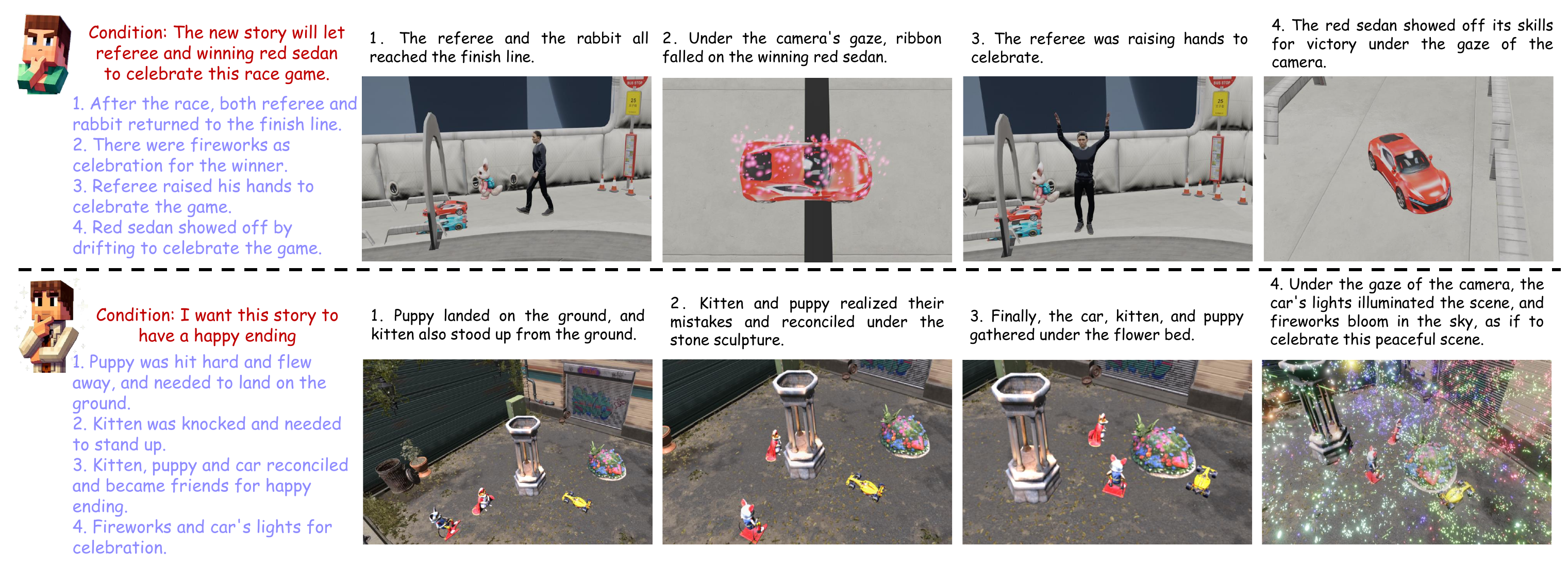}
    \vspace{-8.5mm}
    \caption{We present the story continuation outcomes generated by our \textbf{Story3D-Agent}. Leveraging the rigorous logical reasoning capabilities of LLMs, the newly generated narratives could: 1) Preserve the coherence and consistency of the contextual story content. 2) Align with the stipulated conditions for the new narrative. 3) Implement all generated plots without introducing any misleading elements. The continuation results for the narratives \textit{Race Day} and \textit{Friendship} are provided for illustration.}

    \vspace{-6mm}
    \label{fig:story_continuation}
\end{figure*}

\subsection{Ablation Study}
\label{sub:ablation_study}
\subsubsection{Different Architectural Designs of Story3D-Agent.} 
We conduct ablation on the architectural design of Story3D-Agent. 
Table.~\ref{table:ablation_structure} presents the quantitative metrics of different architectures of the LLM-agents based on narratives \textit{Race Day} and \textit{Friendship}. The results indicate that our Story3D-Agent significantly outperforms other designs under all evaluated criteria.
Specifically, our Story3D-Agent outperforms: 
1) Merging action, motion and decoration agents as a whole, which faces confusion in semantic understanding. 
2) Plain action agent without a hierarchical structure design, which complicates the control of multi-character actions. 
3) Framework without the textual/visual self-check mechanism, as the absence of potential errors identification and reconsideration with extra guidance can make LLMs uninterpretable and lead to more reasoning errors. 
Certain studies~\cite{wang2023voyager, shinn2023reflexion} have also investigated the efficacy of feedback and correction methods. 
According to extensive evaluation, we believe our carefully designed framework has significant superiority in modeling 3D storytelling visualization.

\subsubsection{Different LLMs Base Model.} Furthermore, we compare the performance of open-source LLMs in Table.~\ref{table:ablation_LLMs}. When compared with GPT-3.5-turbo~\cite{chatgpt} and Gemini~\cite{team2023gemini}, using GPT4~\cite{achiam2023gpt} as the base model for LLMs significantly outperforms other results and achieves the best outcomes.

\vspace{+1mm}

\begin{table}[h]
\centering
\resizebox{0.4\textwidth}{!}{%
\begin{tabular}{c|c|c|c|c}
\hline
Exp Name & Ins-Align & CLIP-T & ROUGE-L & BERT \\
\hline
Merge & 0.75 & 22.591 & 0.196 & 0.733 \\
No Hierarchy & 0.8 & 21.80 & 0.241 & 0.741 \\
No Textual self-check & 0.767 & 22.766 & 0.192 & 0.727 \\
No Visual self-check & 0.915 & 22.808 & 0.316 & 0.757 \\
\hline
Ours & 0.933 & 23.081 & 0.322 & 0.771 \\

\hline
\end{tabular}%
}
\vspace{-3mm}
\caption{Ablation on different architectures.}
\label{table:ablation_structure}
\end{table}

\begin{table}[h]
\centering
\resizebox{0.4\textwidth}{!}{%
\begin{tabular}{c|c|c|c|c}
\hline
Exp Name & Ins-Align & CLIP-T & ROUGE-L & BERT \\
\hline
GPT-3.5~\cite{chatgpt} & 0.75 & 20.958 & 0.255 & 0.750 \\
Gemini~\cite{team2023gemini} & 0.792 & 22.318 & 0.292 & 0.752 \\
\hline
GPT-4~\cite{achiam2023gpt} & 0.933 & 23.081 & 0.322 & 0.771 \\

\hline
\end{tabular}%
}
\vspace{-3mm}
\caption{Ablation on different LLMs base model.}
\label{table:ablation_LLMs}
\end{table}

\vspace{-4mm}
\subsection{Story Continuation Visualization}
We further investigate our work on story continuation. Its objective is to empower the LLMs to logically construct an extension of the provided story under certain conditions, and the newly created narrative can be actualized within the capabilities of Story3D-Agent.

We discover that it is essential for LLMs to extend the narrative based on stringent logical reasoning. When creating new story, we require LLMs to reflect the rationale behind the designed contents, where the requirements assist LLMs for more precise continuation contents. In Figure.~\ref{fig:story_continuation}, we exhibit the story continuation visualization of the narratives \textit{Race Day} and \textit{Friendship}. Meanwhile, we also display logical reasoning results generated by LLMs.
For instance, in the story \textit{Friendship}, we customize that the condition is that \texttt{the new story will have a happy ending.} 
The newly generated fragments illustrate: 1) The coherence and consistency of the contextual story content (e.g., kitten and puppy both return to their original states before proceeding plots). 
2) The correspondence of the given condition \texttt{happy ending} (e.g., three characters finally gather under the stone sculpture with celebration of fireworks). 
3) All generated plots can be implemented without producing misleading plots. 
Additionally, similar reasoning behavior are also observed in the story \textit{Race Day} based on the offered condition requires that \texttt{the new story will involve referee and winning red sedan to celebrate this race game}. 
We hope that our work could further advance the practicality of actual 3D visual story creation.

%% file: RealSec1/5_conclusion_v1.tex
\vspace{-3mm}
\section{Conclusion}
\label{sec:Conclusion}

\vspace{-1mm}
In this study, we introduce \textbf{Story3D-Agent}, a novel LLM-agents framework developed for the purpose of modeling 3D storytelling visualization. 
Our approach hinges on the strategic deployment of LLMs via procedural modeling, which effectively facilitates the integration of a multitude of diverse elements, crucial for the delivery of comprehensive visual narrative. 
By leveraging procedural modeling, it becomes possible to execute precise controls for long-range and dynamic 3D representation. 
Moreover, our method could further extends the story through logical reasoning to ensure that the produced content complies cohesively with predefined conditions. 
Through extensive evaluations, we have validated the effectiveness of our framework. We believe 
our innovative method will advance 3D visual content creation in new perspective.

%% file: RealSec1/X_Supp_v1.tex
\vspace{-3mm}
\section{Supplementary Materials of Story3D-Agent}
\label{sec:Supplementary}

\noindent The supplementary document is organized as follows: 
\begin{enumerate}
    \item Details of constructed Python platform. 
    
    \item Implementation Details of Story3D-Agent.
    
    \item Demonstration of story continuation.
    
    \item Explanation of evaluation criteria.
    
    \item Terminology Explanation.

    \item Discussions and Limitations
    
\end{enumerate}

\section{Details of constructed Python platform}
\label{sec:python_platform}
To effectively illustrate 3D narratives incorporating these mentioned factors, we have established a platform comprising three pre-defined Python function libraries, specifically: action, motion, and decoration libraries. 
The guiding principle behind the development of this platform is to cater to as much diversity and creativity in visual story presentation as possible. The subsequent contents will delve into the functions contained within this platform.
We believe that building a basic platform is crucial, which would further advance the 3D visual content creation.

\subsection{Action Library}
The action library consists of Python functions, which aims at controlling actions for characters in story. 
The action library contains seven major action categories, including special action, straight line movement, curved movement, jumping motion, impact motion, state recovery action and demonstration action. For each major category, there will be one or several sub-actions. 

1. special action: there are special actions for characters. In this category, there is one sub-action, namely \texttt{do nothing}.
\begin{itemize}
    \item do nothing: the character will stay in place and do nothing.
\end{itemize}

2. straight line movement: the character will move along the straight line. In this category, there are two sub-actions, namely \texttt{constant speed movement} and \texttt{variable speed movement}.
\begin{itemize}
    \item constant speed movement: the character will move through the straight line at a constant speed.
    \item variable speed movement: the character will move through the straight line at a variable speed.
\end{itemize}

3. curved movement: the character will move along the curve. In this category, there are three sub-actions, namely \texttt{bezier curve movement}, \texttt{S-curve movement} and \texttt{B-curve movement}.
\begin{itemize}
    \item bezier curve movement: the character will conduct curved movement with one control point, and the extent of the curved movement is intense.
    \item S-curve movement: the character will conduct curved movement with two control points, and the extent of the curved movement is moderate.
    \item B-curve movement: the character will conduct curved movement with two control points, and the extent of the curved movement is strange.
\end{itemize}

4. jumping motion: the character will perform jumping motion. In this category, there are two sub-actions, namely \texttt{jump in place} and \texttt{jump forward}.
\begin{itemize}
    \item jump in place: the character will jump in place and the position does not change.
    \item jump forward: the character will jump forward to another position.
\end{itemize}

5. impact motion: the character will move after being collided. In this category, there are three sub-actions, namely \texttt{fall down}, \texttt{knocked down} and \texttt{knocked away}.
\begin{itemize}
    \item fall down: the character will fall to the ground.
    \item knocked down: the character will be heavily knocked to the ground.
    \item knocked away: the character will be very intensely knocked into the air.
\end{itemize}

6. state recovery action: the character will return to its original state. In this category, there are two sub-actions, namely \texttt{stand up} and \texttt{landing from the sky}.
\begin{itemize}
    \item stand up: the character will change from falling to standing up.
    \item landing from the sky: the character will fall from the air to the ground.
\end{itemize}

7. demonstration action: the character will demonstrate its status and appearance. In this category, there are two sub-actions, namely \texttt{rotate in place} and \texttt{drifting}.
\begin{itemize}
    \item rotate in place: the character will continuously rotate left and right in place.
    \item drifting: the character will conduct drifting motion to show winning.
\end{itemize}

\subsection{Motion Library}
The motion library consists of Python functions, which especially aims at controlling motions for humanoid characters in story. We incorporate PriorMDM~\cite{shafir2023human} and ProgMoGen~\cite{liu2024programmable} based on motion diffusion model~\cite{zhang2022motiondiffuse} as the motion controller.
The motion library mainly involves five decoration categories, including special motion, trajectory based generation, human-scene interaction, human-object interaction and physics-based generation.

1. special motion: the humanoid character will stay in place and do nothing.

2. trajectory-based generation: the humanoid character will move along the given trajectory.

3. human-scene interaction: the humanoid character will only walk inside the given square.

4. human-object interaction: the humanoid character will control one object from one place to another.

5. physics-based generation: the humanoid character will stand on one foot and keep balanced.

\subsection{Decoration Library}
The decoration library consists of Python functions, which aims at generating decorations for story and make it more interesting. 
The decoration library contains seven decoration categories, including switching camera perspective, light illumination, particle floc, beaming material, firework, sun light and camera ring shot.

1. switching camera perspective: switch the original camera perspective to the new camera perspective and focus on the current character.

2. light illumination: the light will shine on this character.

3. particle floc: ribbon-like particle floc will be created for decoration on the character.

4. beaming material: the character will emit colored light.

5. firework: creating some fireworks on the ground for celebration.

6. sun light: sun rays dimming or the light in the environment dims.

7. camera ring shot: the camera will take a 360-degree shot around the specified object.

\section{Implementation Details of Story3D-Agent}
\label{sec:implementation_details}
In this section, we provide a comprehensive description of the implementation of the Story3D-Agent. The framework primarily involves four processes: decision-making, textual self-check, execution and visual self-check, and the 3D rendering processes will be facilitated in Blender. 
We will elucidate these processes sequentially, beginning with the roles of the director, action, motion, and decoration agents in the decision-making process. 
Subsequently, we will explain the functioning of the textual self-check mechanism that consists of self-reflection and correction. 
In addition, we will introduce the integration of the execution agent.
Finally, we will introduce the visual self-check for the rendered results. 
For a clear illustration, we will demonstrate these processes using \texttt{Event Window 5} from the story \textit{Race Day} as the example. 
We will reuse the previous symbolic meanings and the pre-defined functions introduced in the \texttt{preliminary of procedural modeling} for further clarification.

\subsection{Workflow of director agent}
\label{sub:workflow_director}
\begin{figure*}[htbp]
\centering
\includegraphics[scale=0.35]{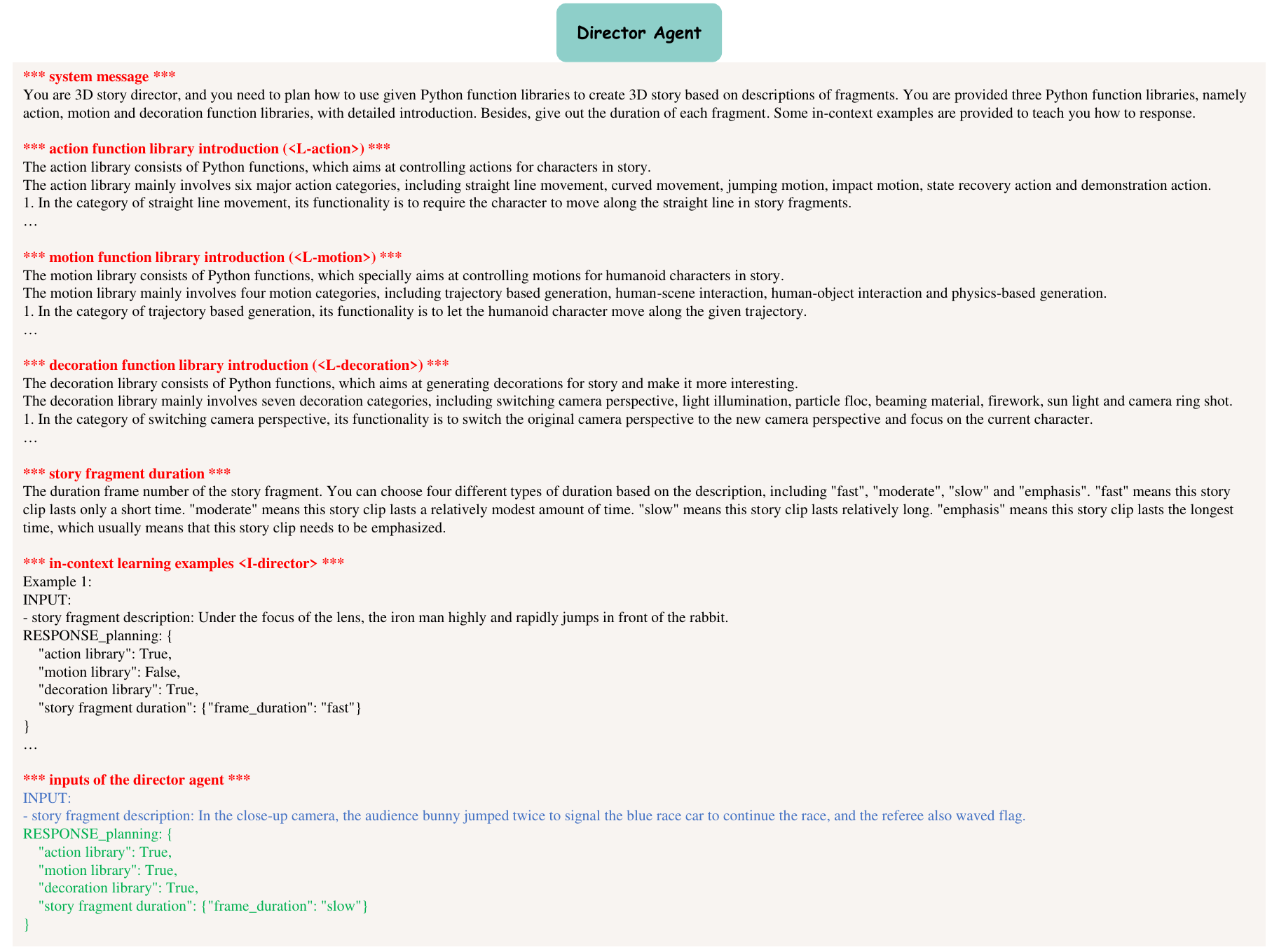}
\vspace{-3mm}
\caption{The workflow of director agent for \texttt{Event Window 5} in \textit{Race Day}.}

\vspace{-15mm}
\label{fig:DirectorAgent_supp}
\end{figure*}

The functionality of the director agent is to ascertain which library or libraries ought to be dispatched and the time duration necessary for the current story fragment. 
Consequently, it receives a comprehensive overview of the action, motion and decoration function libraries, as well as information on time duration types, including "fast", "moderate", "slow", and "emphasis". 
Figure.~\ref{fig:DirectorAgent_supp} illustrates the workflow of the director agent in this scenario. In the given example, all libraries are required (indicated by a \texttt{True} response), and a rather long duration (indicated by a \texttt{slow} response), signifying a relatively extended period, is selected.

\subsection{Workflow of action agent}
\label{sub:workflow_action}
\begin{figure*}[htbp]
\centering
\includegraphics[scale=0.35]{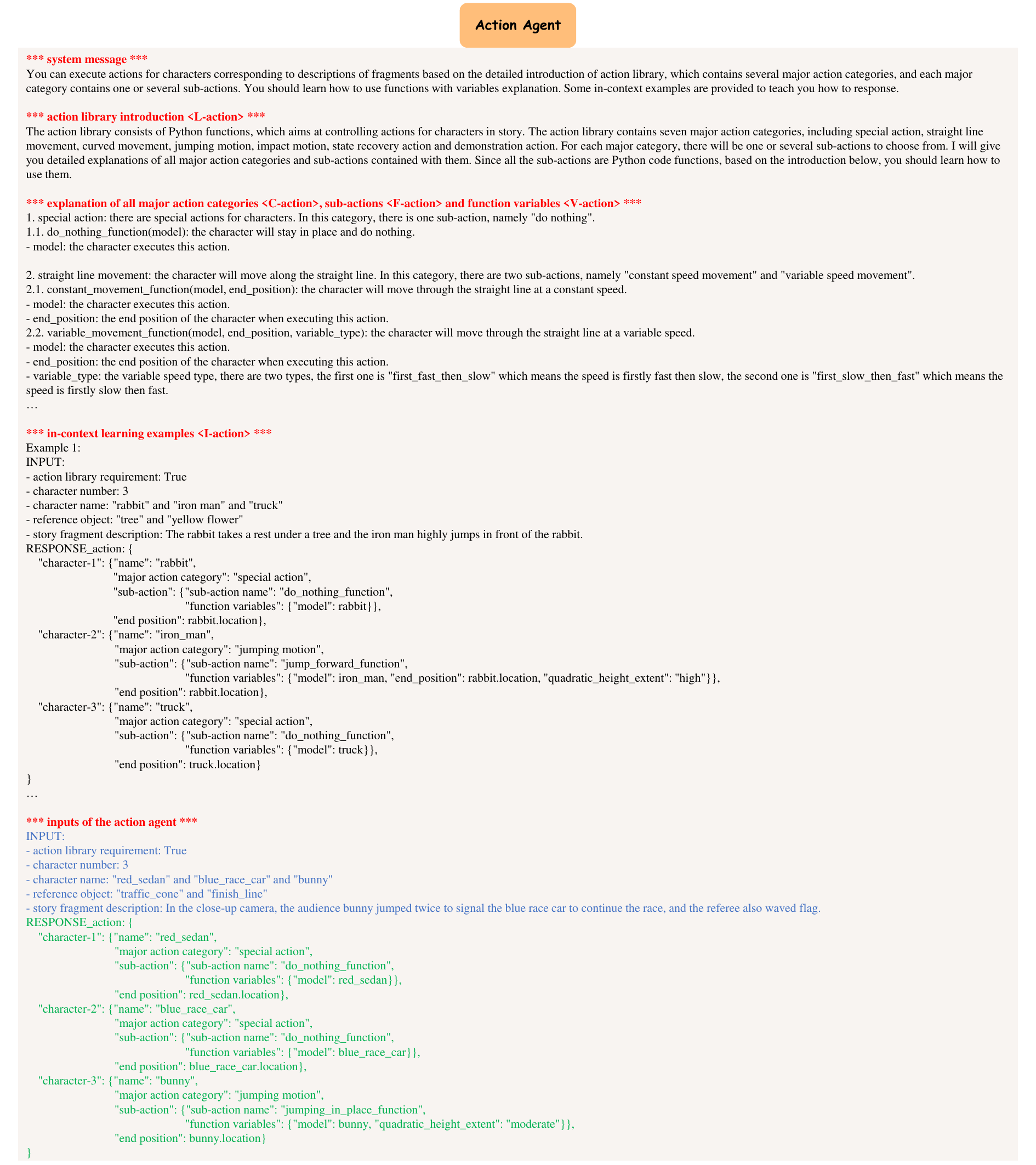}
\vspace{-3mm}
\caption{The workflow of action agent for \texttt{Event Window 5} in \textit{Race Day}.}
\vspace{-5mm}
\label{fig:ActionAgent_supp}
\end{figure*}

The primary responsibility of the action agent is to schedule actions for characters (not including humanoid character) in accordance with the storyline, utilizing the action function library. 
Consequently, it receives a comprehensive overview of the action function library, as well as explanations of the major action categories, sub-actions, and function variables. 
Figure.~\ref{fig:ActionAgent_supp} depicts the workflow of the action agent in this scenario. In the given example, the action agent hierarchically dispatches a \texttt{jumping motion} for the audience bunny initially, followed by a \texttt{jump in place} action for twice. Additionally, it selects the \texttt{do nothing} function in special action for both read sedan and blue race car, as there is no specific action requirement for them in this segment.

\subsection{Workflow of motion agent}
\label{sub:workflow_motion}
\begin{figure*}[htbp]
\centering
\includegraphics[scale=0.66]{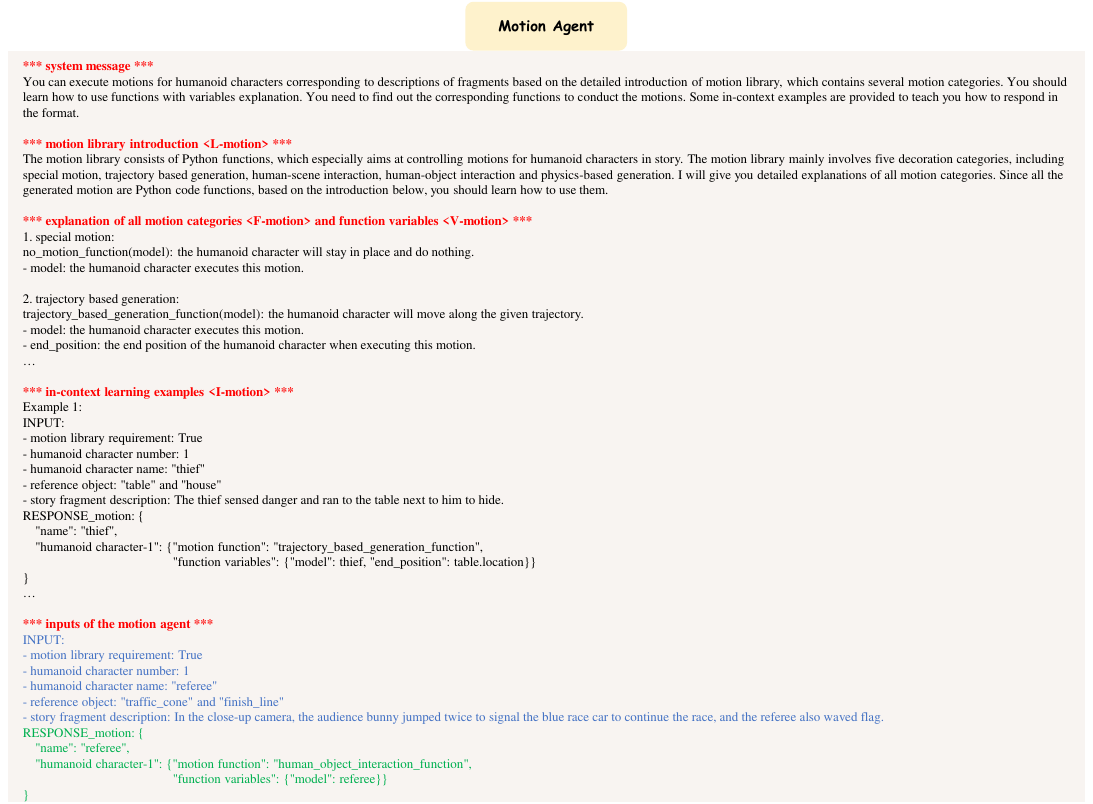}
\vspace{-1mm}
\caption{The workflow of motion agent for \texttt{Event Window 5} in \textit{Race Day}.}

\vspace{-15mm}
\label{fig:MotionAgent_supp}
\end{figure*}

The aim of the motion agent is to specially choose appropriate motions for humanoid characters corresponding to the description, exploiting the motion function library. 
It then receives a comprehensive overview of the motion function library, which includes several different human motion types. Figure.~\ref{fig:MotionAgent_supp} illustrates the operational flow of the motion agent in this clip.
In this example, the motion agent selects the \texttt{human-object interaction} motion function to demonstrate the referee that waving flag.

\subsection{Workflow of decoration agent}
\label{sub:workflow_decoration}
\begin{figure*}[htbp]
\centering
\includegraphics[scale=0.66]{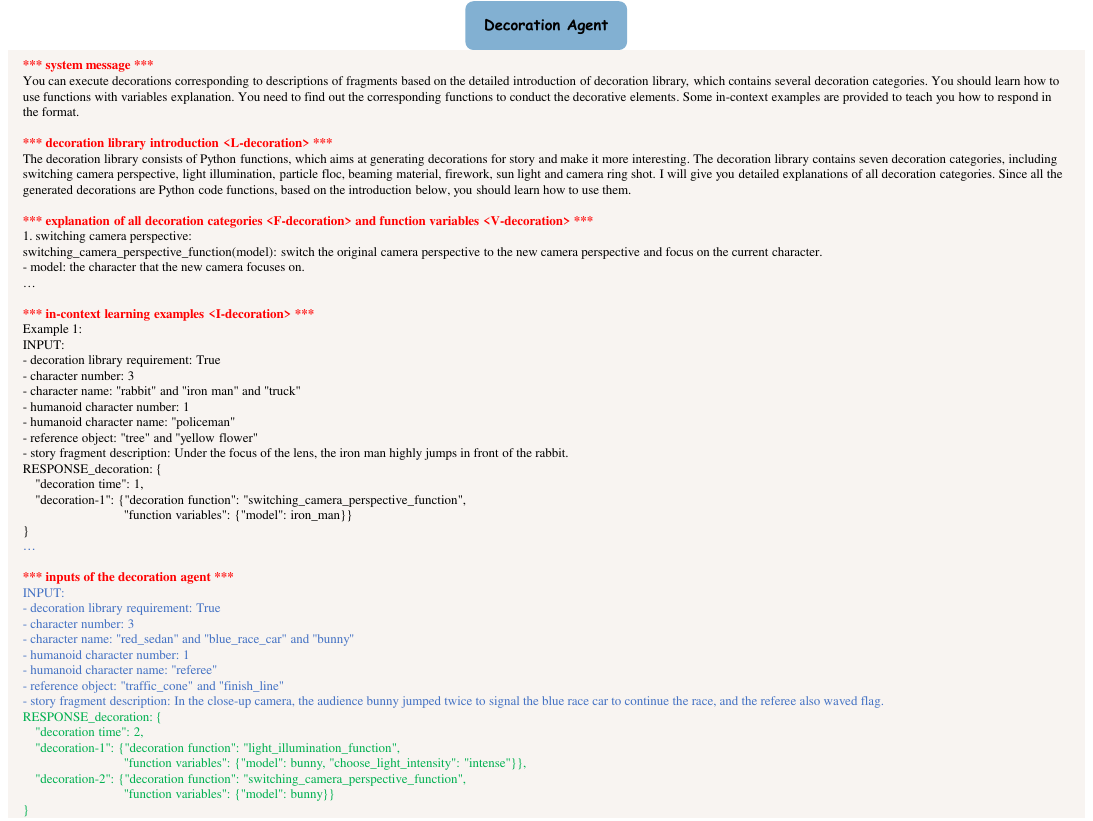}
\vspace{-1mm}
\caption{The workflow of decoration agent for \texttt{Event Window 5} in \textit{Race Day}. It is noteworthy that when the output is judged incorrect, it will undergo the processes of self-reflection and correction.}
\vspace{-5mm}
\label{fig:DecorationAgent_supp}
\end{figure*}

The aim of the decoration agent is to orchestrate decorative elements that are congruent with the narrative within the decoration function library. Subsequently, it is furnished with a comprehensive overview of the decoration function library, inclusive of the categories and function variables. Figure.~\ref{fig:DecorationAgent_supp} illustrates the operational flow of the decoration agent in this scenario.
In the given instance, the decoration agent deploys \texttt{light illumination} for concentrated illumination and \texttt{switching camera perspective} for a different camera perspective.
However, there is no need for light illumination for any character, necessitating the initiation of self-reflection and correction processes in the textual self-check mechanism.

\subsection{Workflow of textual self-check}
\label{sub:workflow_textual_self_check}
\begin{figure*}[htbp]
\centering
\includegraphics[scale=0.44]{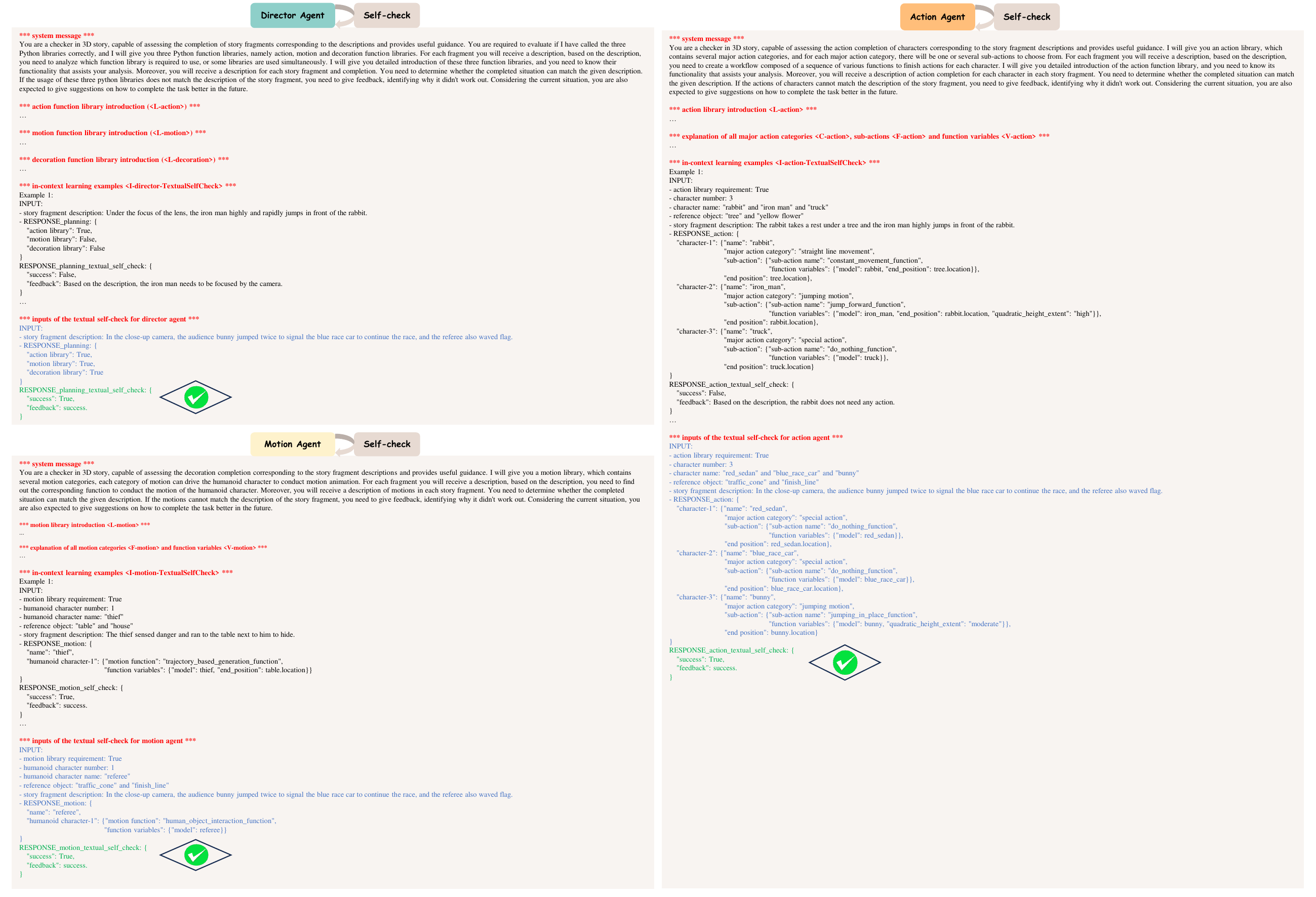}
\vspace{-1mm}
\caption{The workflow of textual self-check mechanism for director, action and motion agents for \texttt{Event Window 5} in \textit{Race Day}.}
\vspace{-4mm}
\label{fig:SelfCheck_2_supp}
\end{figure*}

\begin{figure*}[htbp]
\centering
\includegraphics[scale=0.46]{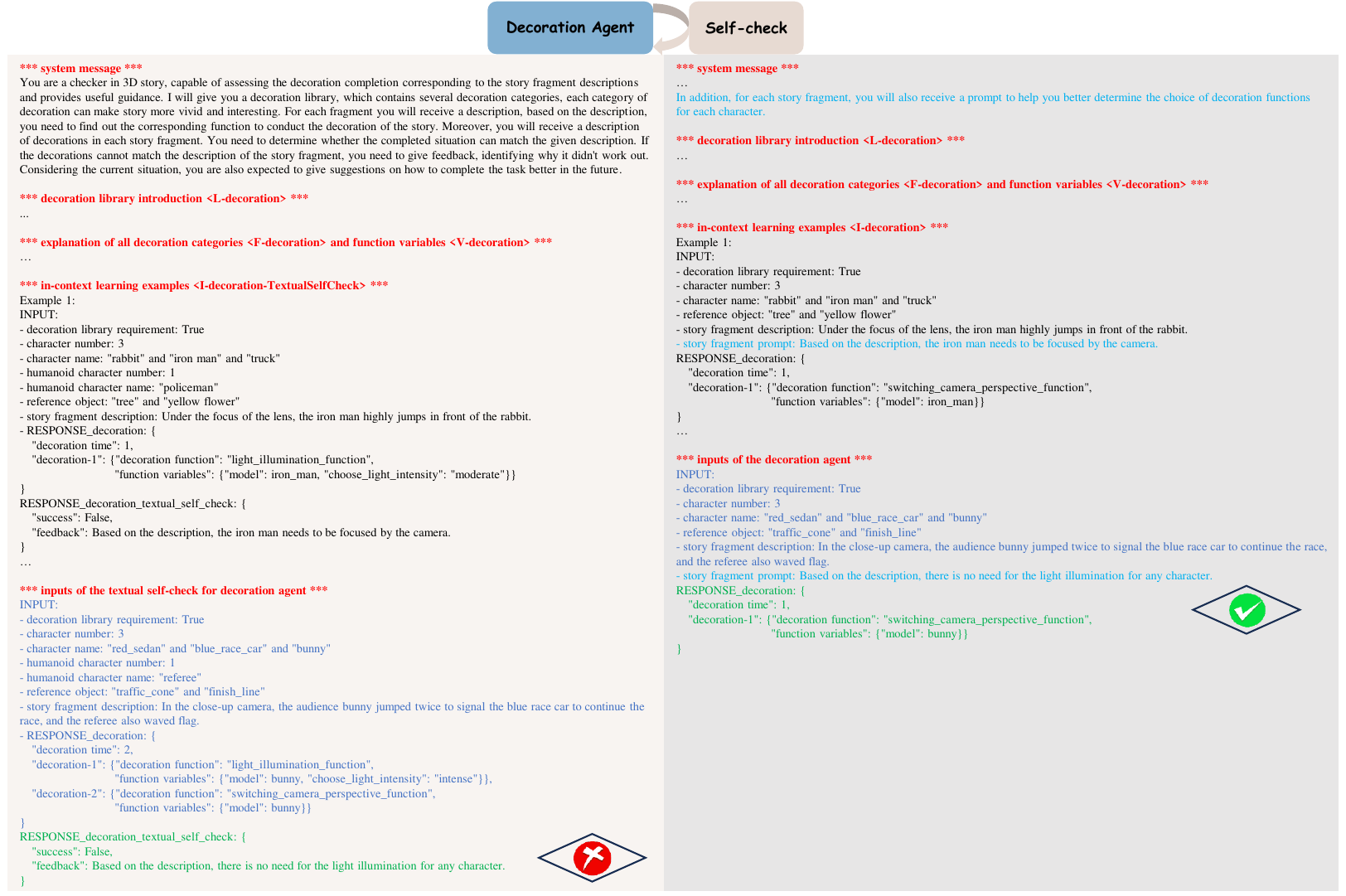}
\caption{The workflow of textual self-check mechanism for decoration agent, the processes of self-reflection and correction are triggered until error-free.}
\vspace{-3mm}
\label{fig:SelfCheck_1_supp}
\end{figure*}

Textual self-check mechanism involves the processes of self-reflection and correction. 
The self-reflection process is designed to ascertain the accuracy of the output results generated by the four aforementioned agents, and to provide feedback when these results are found to be erroneous. 
The correction process then re-initiates the original agents, using the feedback as an additional input prompt. 
The director agent is responsible for assessing the suitability of the library dispatch, while the action, motion and decoration agents are tasked with evaluating the alignment of the selected execution functions with the story plot. 
This process is repeated until the textual self-check mechanism confirms the accuracy of each output result. Figure.~\ref{fig:SelfCheck_2_supp} and Figure.~\ref{fig:SelfCheck_1_supp} depict the workflows of the textual self-check mechanism for these four agents. 
The outputs of the director, motion and action agents are positively confirmed, whereas the textual self-check mechanism within the decoration agent triggers feedback due to potential inaccuracies in responses. 
This necessitates the initiation of the self-reflection process to provide feedback (\eg, \texttt{no need for the light illumination}). Subsequently, the decoration agent is re-executed to rectify the error using the provided feedback. Ultimately, the revised response is deemed correct, marking the completion of the current event window.

Several studies~\cite{yao2022react, shinn2023reflexion, wang2023voyager} have also explored the effectiveness of feedback and correction methods. In the absence of suitable feedback, LLMs may lack crucial judgment about the results of their decisions, potentially resulting in the repeated implementation of unsuccessful strategies. Therefore, the textual self-check mechanism that we propose holds significant importance.

\subsection{Workflow of execution agent}
\label{sub:workflow_execution}
\begin{figure*}[htbp]
\centering
\includegraphics[scale=0.34]{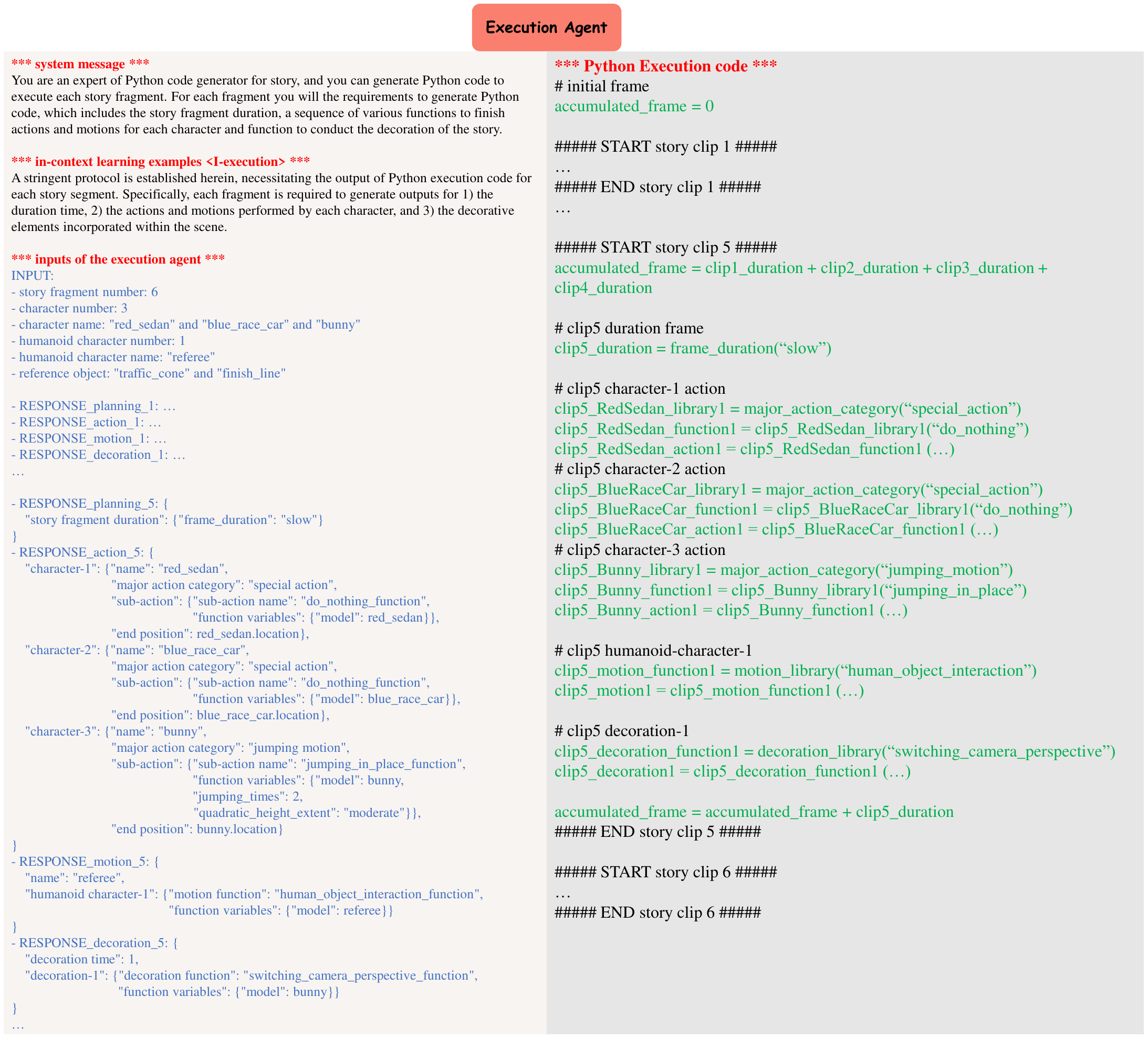}
\caption{The workflow of the execution agent, which amalgamates responses from LLMs and transforms them into executable codes for Blender.}
\vspace{-3mm}
\label{fig:ExecutionAgent_supp}
\end{figure*}

The primary role of execution agent is to seamlessly amalgamate the signals generated by the action, motion and decoration agents, along with the time duration given by the director agent, into executable codes. 
These codes can subsequently be implemented within the software. The execution process is ingeniously designed to adaptively accommodate a variable number of characters, reference objects, and fragments in 3D rendered results. 
This adaptability is particularly advantageous in long-range and dynamic temporal modeling scenarios, as it significantly reduces the human workload and enhances the efficiency of actual visual story content creation.
Figure.~\ref{fig:ExecutionAgent_supp} illustrates the workflow of the execution agent. One particular note is the variable "\texttt{accumulated\_frame}", which ensures that all fragments could strictly adhere to the predetermined storyline, as it is continuously accumulated along the timeline.

\subsection{Workflow of visual self-check}
\label{sub:workflow_visual_self_check}
Visual self-check process involves the examination and assessment of generated visual content, carried out either by human reviewers or Vision-Language Models (VLMs). 
Unlike its textual counterparts, this process evaluates the generated content at the visual level. Once the visual results are validated, the overall system completes its task, or the system issues an additional prompt for the original process to be re-executed, thereby ensuring quality control and accuracy.
The total process is simple, for VLMs, we exploit GPT-4o~\cite{OpenAI2024} and give the prompt as \texttt{Please briefly describe this image in shorts, do not exceed 30 words} to caption the visual results in each fragment, and evaluate if the caption is semantically different from the original description of the fragment.
For humans, they could directly give out the guidance prompt as the additional input for the LLM-agents system.

\section{Demonstration of story continuation}
\label{sub:demonstration_story_continuation}
\begin{figure*}[htbp]
\centering
\includegraphics[scale=0.38]{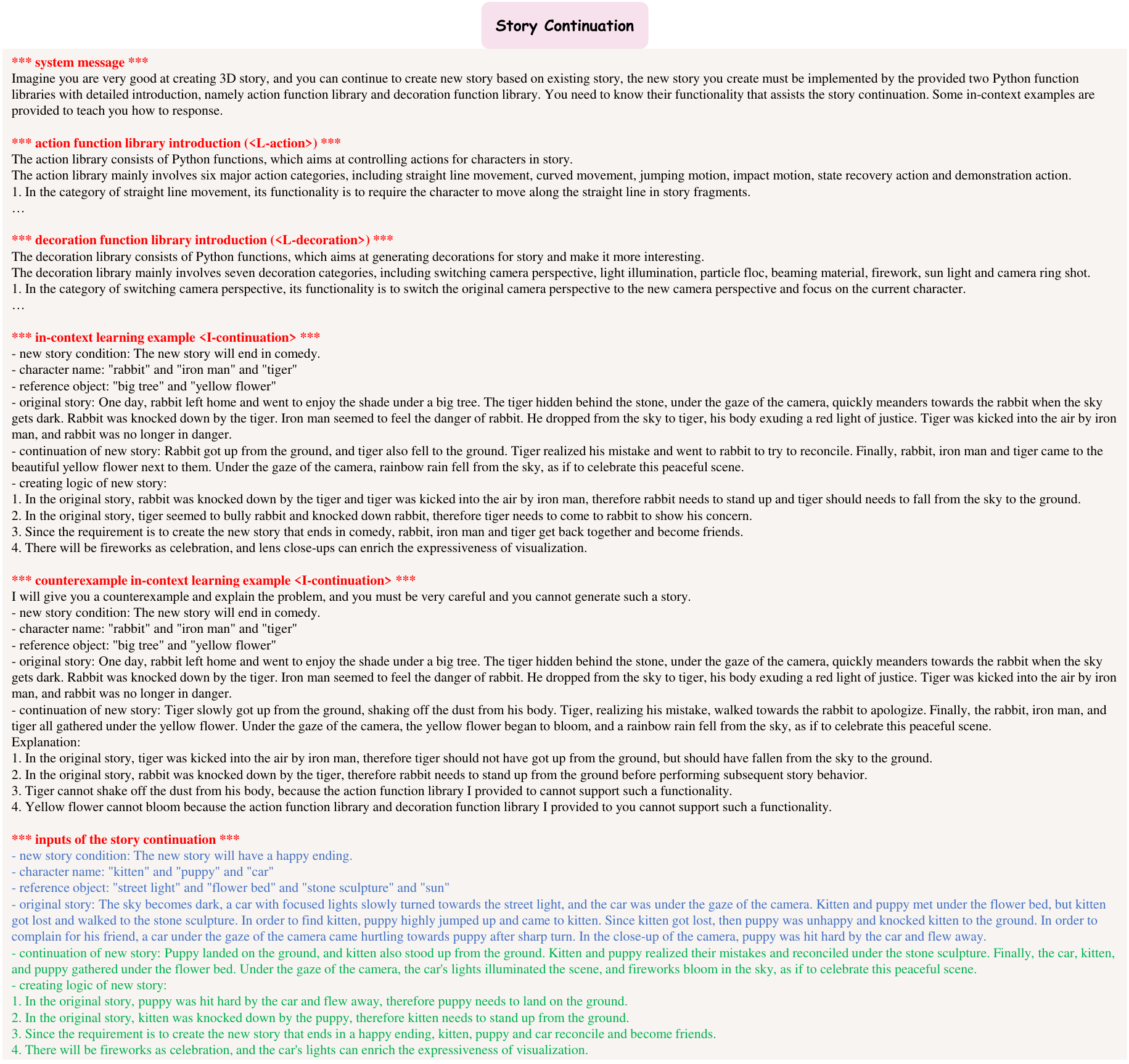}
\caption{The story continuation outcomes of \textit{Friendship}.}
\vspace{-3mm}
\label{fig:StoryContinuation_supp}
\end{figure*}

We further explore the LLMs in enhancing story continuation capability.
When generating new narratives, it is imperative that the created content maintains a consistent and relevant connection to the previously provided materials. 
To meet this criterion, we must: 1) Ensure that the newly created story can be executed by the pre-defined function libraries. 2) Establish the developmental context of the new narrative through specific conditions. 3) Demand that LLMs demonstrate rigorous logical reasoning when crafting new stories, thereby facilitating a more seamless integration of the contextual status and content of the existing narratives. 
Figure.~\ref{fig:StoryContinuation_supp} provides an illustration of the story continuation of \textit{Friendship}. It is noteworthy that we provide not only a common in-context example but also a counterexample. The results indicate that LLMs are capable of creating the required story and demonstrating rigorous logical reasoning for explainable reasons.

\section{Explanation of evaluation criteria}
To assess the accuracy of the alignment between the 3D visualization and the provided narrative, we propose three evaluation criteria: 
1) Instruction-Alignment (Ins-Align): This component employs twenty evaluators who manually assess the concurrence between each rendered segment and the given narrative - ensuring that the visual representation accurately mirrors the story's descriptions.
2) CLIP Score: This metric is computed by calculating the similarities between the depicted results and the original narrative, providing an assessment of the visual alignment with the story.
3) Vocabulary and Semantic-Level Scores: These scores evaluate the captions of rendered results and the original narratives. The process harnesses GPT-4o's~\cite{OpenAI2024} visual understanding capabilities to initially caption the rendered results. Following this, the textual scores are computed using conventional Natural Language Generation (NLG) metrics such as ROUGE-L~\cite{lin2004rouge}, and semantic similarity analysis with the aid of the BERT-base model~\cite{devlin2018bert}.

While the metrics outlined in the main body are evaluated based on two stories, we are not limited to them. To effect a more comprehensive evaluation, we have constructed a test set consisting of 30 story fragments.
We deploy our Story3D-Agent to transcribe these story fragments into 3D rendering results and leverage the previously mentioned three criteria to evaluate the different proposed architectures accordingly. This approach ensures a robust and wide-ranging assessment of our model's performance.
Table.~\ref{table:ablation_structure_supp} indicates that our Story3D-Agent significantly outperforms other designs under all evaluated criteria.

\begin{table}[h]
\centering
\resizebox{0.48\textwidth}{!}{%
\begin{tabular}{c|c|c|c|c}
\hline
Exp Name & Ins-Align & CLIP-T & ROUGE-L & BERT \\
\hline

Merge & 0.728 & 22.609 & 0.201 & 0.723 \\
No Hierarchy & 0.778 & 21.701 & 0.287 & 0.741 \\
No Textual self-check & 0.745 & 22.688 & 0.225 & 0.688 \\
No Visual self-check & 0.893 & 22.853 & 0.284 & 0.721 \\
\hline
Ours & 0.911 & 22.99 & 0.29 & 0.746 \\

\hline
\end{tabular}%
}
\vspace{-3mm}
\caption{Ablation results of 30 story fragments on different architectures.}
\label{table:ablation_structure_supp}
\end{table}

\section{Terminology Explanation}
The concept of \textit{long-range and dynamic} 3D content generation discussed in this paper diverges from the traditional understanding of dynamic generation in 3D creation (e.g., MAV3D~\cite{singer2023text}). Our primary objective is to underscore the distinction from 2D image~\cite{gong2023talecrafter, wang2023autostory} or video-based~\cite{he2023animate, zhuang2024vlogger} storytelling visualization. Previous methods struggle to maintain long-range narrative coherence, and the continuity between the foreground and background across story fragments is often unsatisfactory. Moreover, video-based storytelling visualization is limited in its capacity for substantial transformation and tends towards static (\eg, images) or quasi-static (\eg, videos) representation, which undermines the consistency of dynamic visualization. 
Consequently, we introduce the concept of \textit{long-range and dynamic} content generation in this paper, specifically for 3D story visualization, to highlight this issue.

\section{Discussions and Limitations}

Our Story3D-Agent has shown promising results in the production of 3D storytelling visualization that align closely with the provided narrative. 
However, it is crucial to acknowledge several limitations: 
1) The storytelling visualization expertise is limited, and the non-professional construction of Blender code has resulted in a restricted set of functionalities within the three predefined Python function libraries. Consequently, the complexity of 3D visualization is constrained to some extent. 
2) The quality of generated 3D assets (textures and meshes of characters and reference objects) and human motion is limited by the original model performance. Moreover, more advanced human motion synthesis model, such as whole-body human motion generation~\cite{lin2024motion, lu2023humantomato} and physics-based character control~\cite{zhu2023neural} could better enhance the visual effects of humanoid characters.
3) More professional simulation software (potential higher learning cost), such as Unity\footnote{https://unity.com/} with physical engine, could potentially produce more comprehensive 3D visualization when combined with LLM-agents.

